\documentclass{article} 
\usepackage{main,times}

\usepackage{amsmath,amsfonts,bm}

\def\eqref#1{equation~\ref{#1}}

\def\1{\bm{1}}

\DeclareMathAlphabet{\mathsfit}{\encodingdefault}{\sfdefault}{m}{sl}
\SetMathAlphabet{\mathsfit}{bold}{\encodingdefault}{\sfdefault}{bx}{n}

\usepackage{hyperref}
\usepackage{url}
\usepackage{graphicx}
\usepackage{enumitem}
\usepackage{booktabs}
\usepackage{multirow}
\usepackage{subcaption}
\usepackage{xcolor}
\usepackage{colortbl}
\usepackage{xspace}

\newcommand{\Method}{\texttt{VideoRover}\xspace}
\newcommand{\Bench}{\texttt{VideoRover-Bench}\xspace}

\title{Thinking Beyond Videos: \\
Unifying Video Reasoning and Deep Research for Open-World Video Agents}

\author{\ignorespaces
    \textbf{Wenqi Liu}$^{1,*}$\quad
    \textbf{Shijie Ma}$^{2,*}$\quad
    \textbf{Yunxiao Wang}$^{1,*}$\quad
    \textbf{Meng Liu}$^{1}$\quad
    \textbf{Qile Su}$^{3}$\quad
    \textbf{Han Liu}$^{4}$\quad\\
    \textbf{Bohan Hou}$^{5}$\quad
    \textbf{Zeyu Wang}$^{1}$\quad
    \textbf{Xuanyu Zheng}$^{6}$\quad
    \textbf{Changyi Liu}$^{6}$\quad
    \textbf{Tianke Zhang}$^{6}$\quad
    \textbf{Haonan Fan}$^{6}$\quad\\
    \textbf{Kaiyu Jiang}$^{6}$\quad
    \textbf{Yingxin Li}$^{6}$\quad
    \textbf{Jiankang Chen}$^{6}$\quad
    \textbf{Xu Wang}$^{6}$\quad
    \textbf{Hongyi Fu}$^{6}$\quad
    \textbf{Jianxiong Wang}$^{6}$\quad\\
    \textbf{Bin Wen}$^{6,\ddagger}$\quad
    \textbf{Tingting Gao}$^{6}$\quad
    \textbf{Han Li}$^{6}$\quad
    \textbf{Jianhua Yin}$^{1}$\quad
    \textbf{Yinwei Wei}$^{1,\dagger}$\quad
    \textbf{Xuemeng Song}$^{7,\dagger}$\quad
    \vspace{0.1cm}\\
$^1$Shandong University \quad
$^2$Institute of Automation, Chinese Academy of Sciences \\
$^3$Beihang University\quad
$^4$City University of Hong Kong\quad
$^5$Nanyang Technological University\quad \\
$^6$Kuaishou Technology\quad
$^7$Southern University of Science and Technology
\vspace{0.1cm}\\
\href{https://liuwq-bit.github.io/VideoRover}{\textcolor{blue!50!black}{\texttt{https://liuwq-bit.github.io/VideoRover}}} \\
}

\newif\ificlrarxiv
\iclrarxivtrue  

\ificlrarxiv
  \iclrfinalcopy
\fi

\begin{document}

\maketitle

\ificlrarxiv
  \fancyhead[L]{Preprint}
  \begingroup
  \renewcommand{\thefootnote}{}
  \footnotetext{
    $^*$Equal contribution,
    $^\dagger$Corresponding author,
    $^\ddagger$Project Leader.
  }
  \endgroup
\fi

\begin{abstract}
Open-world video understanding often requires a model to locate sparse visual evidence and acquire external knowledge that is absent from the video and its parametric memory. While Thinking-with-Videos enables active temporal perception and Deep Research supports multi-step information seeking, the two capabilities are typically developed in isolation. We introduce \Method, a unified Video Deep Research framework that iteratively coordinates video cropping, multimodal search, and webpage browsing. Given a video-question pair, \Method uses each tool result to select the next action, so localized video clips guide external retrieval and retrieved evidence triggers further video inspection and verification. To develop this capability, we construct an automated data curation pipeline, producing 26K verified SFT trajectories and 3K challenging RL instances. We also introduce \Bench, a benchmark stratified by video duration and research difficulty. Experiments on VideoDR and \Bench show that our \Method-8B-RL achieves performance comparable to proprietary models in the direct-answer setting without tool use while outperforming larger open-source models equipped with the same tool suite. Ablation studies and training dynamics further validate the complementary roles of active video grounding, external retrieval, and long-horizon reinforcement learning.
\end{abstract}

\section{Introduction}

Real-world multimodal tasks often require both fine-grained visual evidence and open-world knowledge, making them difficult to solve using only the fixed inputs and parametric knowledge of multimodal large language models (MLLMs)~\cite{gao2023retrieval, abootorabi2025ask, xu2025comprehensive, shi2025deep}. Tool-augmented agents address these limitations along two complementary directions. For \emph{active perception}, Thinking-with-Images~\cite{zheng2026deepeyes, zhang2026thyme, shen2026lookwise} allows models to inspect visual regions during reasoning, while Thinking-with-Videos~\cite{meng2026watch, zhang2026dynframe} extends this capability to temporal localization, segment cropping, and adaptive resampling. For \emph{information acquisition}, retrieval-augmented generation (RAG)~\cite{chen2022murag, wang2024searching} introduces external knowledge, while Search~\cite{jin2025search, wu2026mmsearch} and Deep Research~\cite{zheng2025deepresearcher, li2026webthinker} support multi-round text and image search, webpage browsing, and evidence verification. The former determines what evidence to observe from the video, whereas the latter determines what knowledge to acquire beyond it. Open-world tasks grounded in video require both capabilities to work together.

This combination is essential in settings such as news, demonstrations, documentaries, and lectures, where briefly appearing entities or clues distributed across distant segments must be linked to background, historical, or up-to-date information~\cite{liang2025video, liu2026watching}. Thinking-with-Videos can locate the relevant visual cues, but the required knowledge may be absent from the video or may have emerged after it was recorded. External retrieval can supply this knowledge only when grounded in the correct visual anchor. A one-way pipeline of video sampling, query generation, and web search therefore propagates early perception errors and cannot use retrieved evidence to revisit uncertain video content. We argue that \emph{Video Deep Research} should instead allow video observations to guide retrieval and retrieved evidence to refine subsequent video inspection.

Despite progress along both directions, three challenges remain. First, temporal grounding must convert sparse evidence in long videos into reliable anchors for external retrieval. Existing Thinking-with-Videos methods improve evidence access through localization, keyframe selection, and adaptive resampling~\cite{zhang2026thinking, zeng2026video, liu2026videotempo}, but use localized observations primarily for in-video reasoning, without connecting them to external retrieval or revising localization from retrieved evidence. Second, research planning must select among image search, text search, and webpage inspection while formulating effective queries from temporally grounded cues. Existing Deep Research agents mainly operate over text, static images, and webpages~\cite{jin2025search, wu2026mmsearch, huang2026vision}, leaving this cross-modal decision process underexplored. Third, video and retrieval actions must update a shared research state so that each result can redirect subsequent decisions. Yet the two families typically optimize their respective stages separately, and available data rarely cover complete trajectories from temporal grounding and multimodal retrieval to evidence revision and answer synthesis. Consequently, simply attaching search tools to a video reasoner does not provide the coordination required for Video Deep Research.

\begin{figure}[t]
    \centering
    \includegraphics[width=\linewidth]{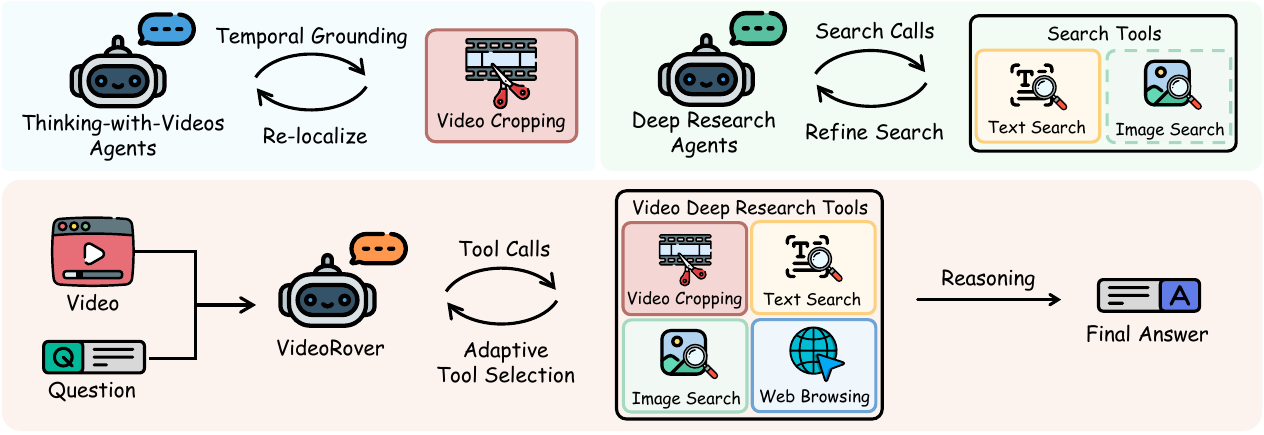}
    \vspace{-15pt}
    \caption{Comparison of Thinking-with-Videos, Deep Research, and \Method. \Method iteratively coordinates video cropping, multimodal search, and webpage browsing, using each result to guide the next action.}
    \label{fig:videorover_main}
    \vspace{-10pt}
\end{figure}

To address these challenges, we introduce \Method, a unified Video Deep Research framework for joint video grounding and multimodal retrieval (Figure~\ref{fig:videorover_main}). For reliable temporal grounding, \Method first localizes and densely inspects a segment, turning sparse video evidence into a visual anchor. For adaptive research planning, it selects a keyframe for image retrieval, formulates text queries for missing knowledge, and visits webpages according to the current \emph{evidence gap}. For iterative coordination, every tool result updates a shared research state and determines whether to continue searching, re-localize or re-examine video evidence, or answer. Video observations thereby ground external search, while retrieved evidence can redirect video reasoning. To provide the required supervision, we develop an automated pipeline that constructs questions jointly dependent on video and external knowledge and synthesizes interaction trajectories, yielding 26K verified SFT examples and 3K challenging RL instances. We further introduce \Bench, stratified by video duration and task difficulty, to evaluate video grounding, open-world retrieval, and multi-source reasoning across different temporal and research complexities.

Our main contributions are summarized as follows:
\begin{itemize}[leftmargin=15pt]
    \item We propose \Method, which unifies Thinking-with-Videos and Deep Research in an agent that uses each tool result to coordinate video observation and open-world retrieval.
    \item We develop an automated pipeline for constructing Video Deep Research questions and verified trajectories involving temporal localization, multimodal retrieval, and evidence synthesis.
    \item We introduce \Bench, a stratified benchmark spanning different video durations, task difficulty levels, and reasoning complexities.
    \item Experiments on VideoDR and \Bench demonstrate strong Video Deep Research performance, while ablation studies and training dynamics validate the complementary tool design and long-horizon coordination learned through RL.
\end{itemize}

\section{Related Work}

\subsection{Thinking-with-Videos}

Recent multimodal large language models (MLLMs) have advanced video understanding and reasoning, yet uniform sampling under a fixed visual token budget can easily miss sparse evidence in long videos. LongVA~\cite{zhang2024long} and LongVILA~\cite{chen2025longvila} increase context capacity, while token compression and hierarchical modeling improve efficiency, but these approaches typically determine the visual input before inference and cannot adapt observation to intermediate reasoning needs. Tool-augmented methods instead treat visual content as an active reasoning workspace: OpenThinkIMG~\cite{su2025openthinkimg}, DeepEyes~\cite{zheng2026deepeyes}, and Thyme~\cite{zhang2026thyme} enable iterative image inspection, while VITAL~\cite{zhang2026thinking}, LongVT~\cite{yang2026longvt}, FrameThinker~\cite{he2026framethinker}, Video-o3\cite{zeng2026video} and VideoTemp-o3~\cite{liu2026videotempo} extend this idea to temporal localization, keyframe selection, and adaptive video resampling. Despite improving access to sparse temporal evidence, existing Thinking-with-Videos methods primarily reason within the input video. \Method further connects localized video evidence to image search and external retrieval, allowing retrieved knowledge to verify and redirect subsequent video reasoning.

\subsection{Deep Research}

Early search agents such as WebGPT~\cite{nakano2021webgpt} and Search-R1~\cite{jin2025search} extend retrieval-augmented generation with iterative reasoning and search, but remain primarily text-centric. Recent multimodal systems further incorporate visual search and interaction: MMSearch-R1~\cite{wu2026mmsearch} and WebWatcher~\cite{geng2026webwatcher} learn multimodal search behaviors through reinforcement learning, while Vision-DeepResearch~\cite{huang2026vision}, MM-DeepResearch~\cite{yao2026mm}, OpenSearch-VL~\cite{chen2026opensearch}, and HyperEyes~\cite{li2026hypereyes} further support long-horizon, multi-tool visual--textual exploration with capabilities such as multi-entity search, visual manipulation, and cross-modal evidence synthesis. These methods substantially extend Deep Research beyond text, yet their perceptual inputs remain largely static images or webpages, without explicitly addressing sparse evidence distributed over long videos. \Method targets this gap by placing temporal localization, image search, text search, and webpage browsing in a unified trajectory, where video evidence grounds external retrieval and retrieved knowledge can trigger video revisiting and verification.


\section{Task Formulation}
\label{sec:task_formulation}

We define a Video Deep Research instance as $x=(V,Q)$, where $V=(v_1,\ldots,v_N)\in\mathcal{V}$ is an $N$-frame video and $Q\in\mathcal{Q}$ is a research question. Given an open-world web environment $\mathcal{W}$, the goal is to produce $Y\in\mathcal{Y}$ by identifying the relevant video evidence and acquiring any missing external knowledge.
At step $t$, we represent the current research state as:
\begin{equation}
    S_t=(G_t^{v},G_t^{w},h_t),
\end{equation}
where $G_t^{v}$ and $G_t^{w}$ are the accumulated video and web evidence, and $h_t$ is the interaction history. The model selects an action $u_t$ from the remaining evidence gap and updates the state with the tool observation $z_t$:
\begin{equation}    
u_t\sim\pi_\theta(\cdot\mid Q,S_t),\qquad
S_{t+1}=\operatorname{Update}(S_t,u_t,z_t).
\end{equation}
\Method uses four tools. The $\operatorname{crop\_video}(V,[s_t,e_t])$ tool returns a densely sampled clip from a selected interval. The $\operatorname{image\_search}(v_{k_t})$ tool retrieves visually related information using a keyframe, while $\operatorname{text\_search}(q_t)$ returns candidate webpages and summaries for a text query. Finally, $\operatorname{visit}(l_t)$ opens a webpage to collect detailed evidence. These tools may be invoked repeatedly until the model executes the terminal action $\operatorname{answer}$.

We denote the complete interaction trajectory as:
\begin{equation}
\tau=\big(S_0,(u_1,z_1),\ldots,(u_T,z_T),Y\big),    
\end{equation}
where $S_0$ contains $x$ and an initial sparse video sample. The process terminates when the accumulated evidence supports an answer or the interaction budget $T_{\max}$ is reached.

\section{Data Synthesis and Benchmark Construction}

High-quality training data is essential for developing Video Deep Research capabilities. Unlike conventional video question answering, Video Deep Research requires not only reasoning over video content, but also acquiring external information beyond the model's parametric knowledge and connecting video evidence with web evidence through multi-step interactions. To support this setting, we develop a complete data pipeline covering Video Deep Research question construction, SFT trajectory synthesis, and hard-example selection for reinforcement learning. We further construct a stratified Video Deep Research benchmark, \Bench, organized by video duration and task difficulty.

\subsection{Question Construction}

\begin{figure}[t]
    \centering
    \includegraphics[width=\linewidth]{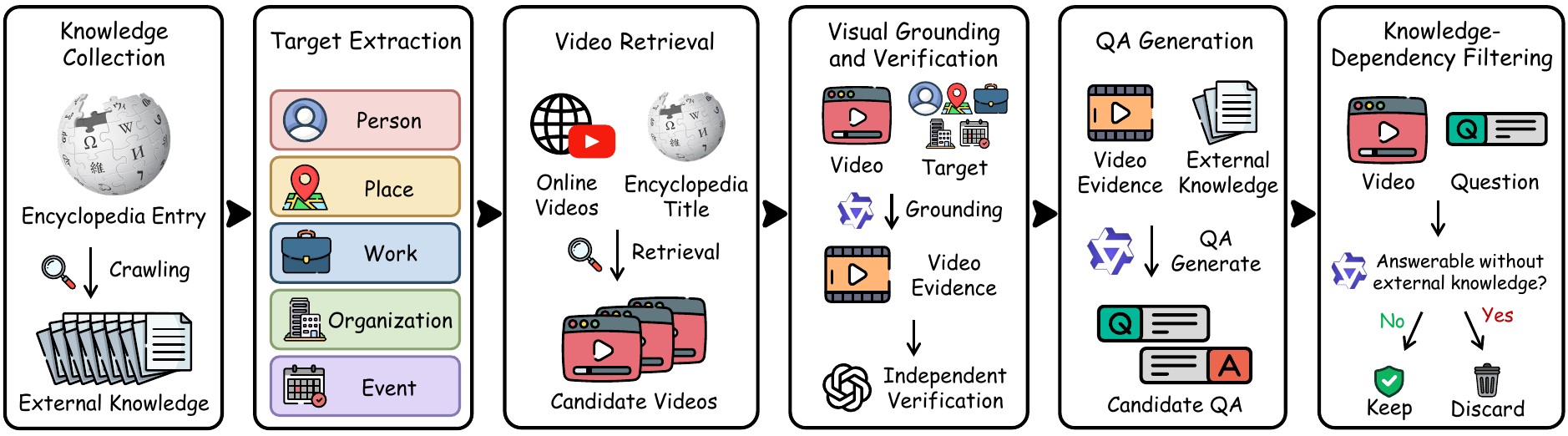}
    \caption{Pipeline for constructing Video Deep Research questions.}
    \label{fig:qa_generation}
\end{figure}

Figure~\ref{fig:qa_generation} summarizes how we construct task instances $x=(V,Q)$ with reference answers $Y$, following Section~\ref{sec:task_formulation}. Each instance enforces two complementary dependencies: the video identifies a target referenced only indirectly by the question, while external web knowledge provides the missing information needed for the answer. Thus, neither video-only reasoning nor question-only web search is sufficient.

\paragraph{Knowledge collection and target extraction.}
We collect Wikipedia articles and extract visually identifiable people, places, organizations, works, products, and events as candidate targets. For each target, we retain the article title, supporting text, and source URL as verifiable evidence for question generation and answer checking, while discarding candidates without clear factual answers or reliable sources.

\paragraph{Video retrieval and visual grounding.}
We query YouTube with each target and its Wikipedia title and retain the top 3 candidate videos. To eliminate metadata-only matches, Qwen3.5-27B~\cite{qwen3.5} analyzes 512 densely sampled frames and subtitles to localize likely intervals and keyframes. GPT-5.4-mini~\cite{singh2025openai} then independently verifies the selected keyframe against the target identity. Only video--target pairs passing both stages are retained as reliable visual anchors for subsequent question answering.

\paragraph{Video-grounded question generation.}
Given a grounded video--target pair and its collected web evidence, we generate $Q$ by referring to the target indirectly (e.g., ``a book cover displayed in the video'') and asking about related background, history, or recent developments. The explicit identity is withheld so that answering the question requires first grounding the target in $V$. We then disable retrieval and ask Qwen3-VL-4B~\cite{bai2025qwen3} to answer using only $V$ and $Q$, removing samples it answers correctly. This filtering ensures that the final answer depends on both video evidence and external retrieval.

\subsection{Trajectory Generation}
\label{sec:trajectory_generation}

\begin{figure}[t]
    \centering
    \includegraphics[width=\linewidth]{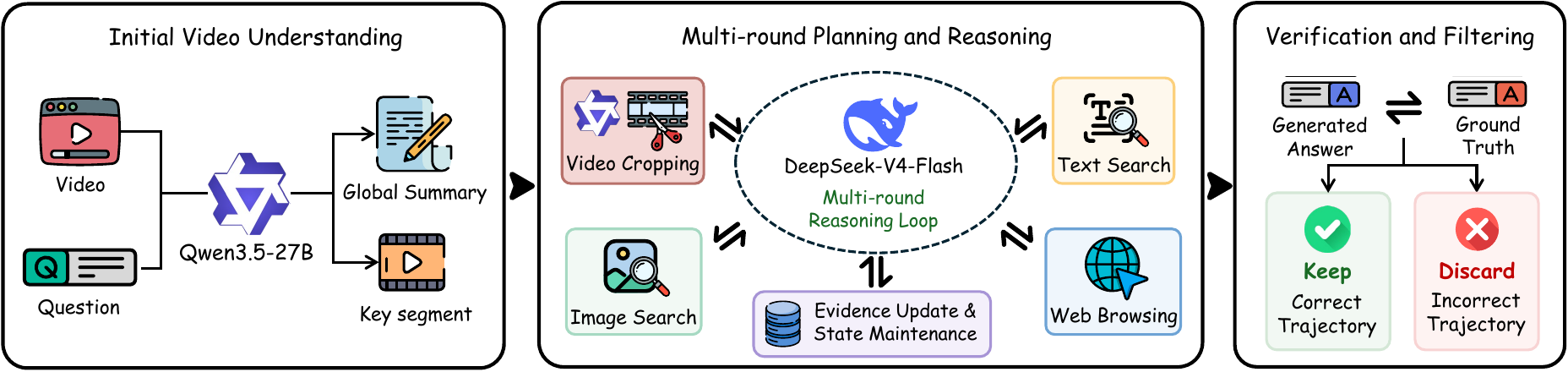}
    \caption{Pipeline for synthesizing Video Deep Research trajectories.}
    \label{fig:trajectory_synthesis}
\end{figure}

High-quality interaction trajectories are essential for teaching the model Video Deep Research behaviors rather than only final-answer generation. As summarized in Figure~\ref{fig:trajectory_synthesis}, our synthesis process explicitly demonstrates how to ground the question in the video, acquire external evidence with multiple tools, and update the research state before answering.

\paragraph{Roles and initialization.}
Each trajectory is jointly generated by DeepSeek-V4-Flash\footnote{The DeepSeek-V4-Flash~ used here is the Preview version released on April 24, 2026.}~\cite{xu2026deepseek}, which serves as the research planner, and Qwen3.5-27B, which serves as the video observer. Because DeepSeek-V4-Flash is a text-only model without visual perception, Qwen3.5-27B inspects the sampled frames and cropped segments and provides temporally grounded visual observations to the planner. The observer first summarizes sparsely sampled frames and proposes an initial relevant interval. Using the same observer as in question construction maintains consistent visual grounding. We require the planner's first response in every trajectory to localize a candidate key segment and invoke \texttt{crop\_video}. The observer then analyzes the densely sampled crop for relevant entities, actions, text, and temporal cues, ensuring that subsequent retrieval is grounded in fine-grained video evidence rather than the initial sparse overview.

\paragraph{Evidence-guided tool use and filtering.}
Subsequent actions are selected from the crop observations and the remaining evidence gap. If the target is still absent, the planner localizes and crops another segment. When a visual entity appears, it prioritizes \texttt{image\_search} to confirm the identity and reduce visual misidentification. Once the entity is identified but the required background, historical, or up-to-date fact remains missing, it invokes \texttt{text\_search}. When either search tool returns a relevant entry, \texttt{visit} opens the source webpage to extract detailed evidence rather than relying only on search-result summaries. Each result updates the research state and can support, reject, or revise existing hypotheses. The loop ends when video and external evidence jointly support an answer, and only trajectories whose final answers match the ground truth are retained.

\subsection{Training Data}
\label{sec:training_data}

\paragraph{SFT Data.}
The synthesis process independently produces answers consistent with the ground truth, providing an additional check of question validity and trajectory correctness. We partition the verified trajectories by interaction length and retain those with at most 10 tool calls for supervised fine-tuning, yielding 26K SFT trajectories. These data cover the fundamental Video Deep Research behaviors of video grounding, multimodal retrieval, evidence updating, and answer synthesis, while longer trajectories are reserved as candidates for subsequent stages.

\paragraph{RL Data.}
After the cold start of SFT, we sample 5 complete rollouts per candidate question whose reference trajectory requires more than 10 tool calls, recording answer correctness, valid tool usage, and interaction length. We remove relatively easy questions answered correctly in more than 3 rollouts, then balance the remaining hard examples by visual-grounding difficulty, search frequency, and overall interaction complexity. This produces 3K RL instances whose rollout groups typically contain both successful and unsuccessful behaviors, providing informative within-group reward variation for relative policy optimization and emphasizing long-horizon autonomous decision making.

\subsection{Benchmark Construction}

\begin{figure}[t]
    \centering
    \includegraphics[width=\linewidth]{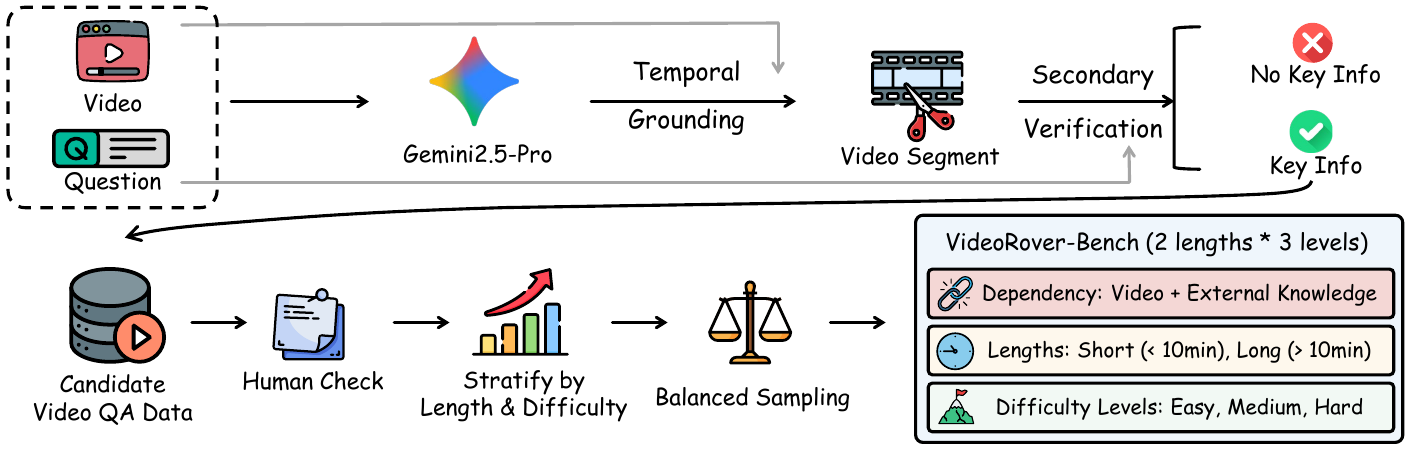}
    \caption{Construction of \Bench.}
    \label{fig:bench_construction}
\end{figure}

We construct \Bench to evaluate Video Deep Research across different temporal scales and research complexities. Figure~\ref{fig:bench_construction} summarizes its verification, stratification, and balanced-sampling pipeline.

Starting from the candidate video--question pool, we use Gemini-2.5-Pro~\cite{comanici2025gemini}, which is independent of the data-generation models, to check question clarity and localize the relevant video segment. The localized segment is then paired with the question for a subsequent verification pass that determines whether it contains the key visual information needed by the question. Samples without such evidence are discarded. The remaining samples undergo human review of question and answer correctness and joint dependence on video evidence and external knowledge. Finally, verified samples are stratified by video length and task difficulty before balanced sampling.

We divide videos into \emph{short} (0--10 minutes) and \emph{long} ($>$10 minutes) subsets, and further categorize questions by temporal-grounding and retrieval difficulty:
\begin{itemize}[leftmargin=15pt]
    \item \textbf{Easy:} the relevant visual evidence is readily localized and only a few searches are required.
    \item \textbf{Medium:} the evidence is more difficult to localize or the answer requires more retrieval steps.
    \item \textbf{Hard:} the visual evidence is elusive, or the answer requires joint reasoning over multiple video segments and search results.
\end{itemize}
Balanced sampling retains 50 examples for each of the 6 length--difficulty combinations, yielding \textbf{300 evaluation instances} in total.

\section{Model Training}

We train \Method in two complementary stages. SFT establishes a reliable tool-use prior that links video observation with external retrieval, while RL improves autonomous decisions under actual tool feedback and final-answer supervision.

\subsection{SFT Training}

We first fine-tune the base model on the verified multi-round trajectories from Section~\ref{sec:trajectory_generation}. Unlike final-answer supervision, these trajectories demonstrate how to localize video evidence, select image or text search, inspect webpages, update the evidence state, and decide whether to continue or answer. SFT therefore provides the model with a stable behavioral initialization for the complete Video Deep Research process, including valid tool invocation and decisions based on returned evidence. Let $x_i=(V_i,Q_i)$ denote the task input and $\tau_i^{*}$ its target trajectory. Tool outputs are treated as environmental context and excluded from the loss over model-generated reasoning, tool calls, and answer tokens. The objective is $\mathcal{L}_{\mathrm{SFT}}(\theta)=-\frac{1}{M}\sum_{i=1}^{M}\log\pi_\theta(\tau_i^{*}\mid x_i)$.

\subsection{RL Training}

Although SFT establishes the basic research behavior, it imitates fixed paths and does not optimize success under actual tool feedback. We therefore apply Group Sequence Policy Optimization (GSPO)~\cite{zheng2025group} to the challenging data from Section~\ref{sec:training_data}, allowing the policy to explore alternative decisions and reinforce complete observation--retrieval--verification trajectories through final-answer correctness. For each input $x$, the old policy samples $G=8$ trajectories. Correct and incorrect answers receive rewards of 1 and 0, while trajectories with formatting errors, missing answers, or invalid tool calls are filtered out. We estimate trajectory-level advantages using the leave-one-out baseline $A_i=R_i-\frac{1}{G-1}\sum_{j\ne i}R_j$.

Let $y_i$ be the $i$-th trajectory and $\rho_i(\theta)$ the sequence-level importance ratio between the current and old policies. GSPO clips this ratio and first averages over valid tokens within each trajectory, then across trajectories:

\begin{equation}
\mathcal{L}_{\mathrm{GSPO}}(\theta)
=-\frac{1}{G}\sum_{i=1}^{G}\frac{1}{|y_i|}\sum_{t=1}^{|y_i|}
\min\!\left(
\rho_i(\theta)A_i,
\operatorname{clip}\!\left(\rho_i(\theta),1-\epsilon_{\mathrm{low}},1+\epsilon_{\mathrm{high}}\right)A_i
\right),
\end{equation}

where $\rho_i(\theta)$ is computed from the mean token-level log-probability ratio within the trajectory. This trajectory-balanced objective prevents longer outputs from dominating optimization. Together with challenging data and outcome-based rewards, GSPO turns the SFT initialization into an adaptive policy that learns when to gather more evidence, revisit the video, or answer.

\section{Experiment}

\subsection{Experimental Setup}

\paragraph{Training Details}
We use Qwen3-VL-8B~\cite{bai2025qwen3} as the backbone of \Method and optimize it through SFT followed by RL. Full implementation and training details are provided in Appendix~\ref{app:experimental_details}.

\paragraph{Evaluation Details}
We evaluate \Method on VideoDR~\cite{liu2026watching} and our proposed \Bench. We compare proprietary models under direct answering and open-source models under both direct answering and ReAct-style agentic tool use~\cite{yao2023react}. All agentic models can autonomously invoke the same tools under a shared evaluation protocol. The complete baseline list and evaluation settings are provided in Appendix~\ref{app:experimental_details}.

\subsection{Main Results}

\begin{table}[t]
    \centering
    \caption{Main results on VideoDR and \Bench. All values are accuracy (\%). S and L denote short and long videos, respectively. Avg. is the unweighted mean over VideoDR and the six \Bench subsets. \Method variants are shaded in blue. Within the open-source agentic group, the best and second-best results are shown in bold and underlined, respectively.}
    \label{tab:main_results}
    \resizebox{\linewidth}{!}{%
    \setlength{\tabcolsep}{4.5pt}
    \renewcommand{\arraystretch}{1.08}
    \begin{tabular}{lcccccccc}
        \toprule
        \multirow{2}{*}{\textbf{Model}} & \multicolumn{1}{c}{\textbf{VideoDR}} & \multicolumn{6}{c}{\textbf{VideoRover-Bench}} & \multirow{2}{*}{\textbf{Avg.}} \\
        \cmidrule(lr){2-2} \cmidrule(lr){3-8}
        & \textbf{Acc.} & \textbf{S-Easy} & \textbf{S-Medium} & \textbf{S-Hard} & \textbf{L-Easy} & \textbf{L-Medium} & \textbf{L-Hard} \\
        \midrule
        \rowcolor{gray!20}
        \multicolumn{9}{c}{\textbf{Proprietary Models (Direct Answer)}} \\
        GPT-5                 & 57.00 & 70.00 & 64.00 & 36.00 & 64.00 & 50.00 & 44.00 & 55.00 \\
        GPT-5.2               & 31.00 & 56.00 & 52.00 & 24.00 & 60.00 & 30.00 & 24.00 & 39.57 \\
        GPT-5.4               & 51.00 & 56.00 & 54.00 & 26.00 & 24.00 & 24.00 & 16.00 & 35.85 \\
        Gemini-2.5-Flash      & 47.00 & 54.00 & 30.00 & 30.00 & 54.00 & 30.00 & 28.00 & 39.00 \\
        Gemini-2.5-Pro        & 54.00 & 66.00 & 54.00 & 30.00 & 62.00 & 34.00 & 50.00 & 50.00 \\
        Gemini-3-Flash        & 59.00 & 74.00 & 70.00 & 48.00 & 58.00 & 52.00 & 48.00 & 58.42 \\
        Gemini-3-Pro          & 55.00 & 78.00 & 60.00 & 42.00 & 72.00 & 40.00 & 52.00 & 57.00 \\
        \midrule
        \rowcolor{gray!20}
        \multicolumn{9}{c}{\textbf{Open-Source Models (Direct Answer)}} \\
        Qwen3-VL-8B           & 9.00  & 8.00  & 8.00  & 8.00  & 18.00 & 6.00  & 10.00 & 9.57  \\
        Qwen3-VL-30B-A3B      & 10.00 & 14.00 & 16.00 & 12.00 & 22.00 & 10.00 & 12.00 & 13.71 \\
        Qwen3.5-27B           & 28.00 & 32.00 & 30.00 & 14.00 & 28.00 & 22.00 & 12.00 & 23.71 \\
        Qwen3.5-35B-A3B       & 24.00 & 36.00 & 28.00 & 14.00 & 36.00 & 20.00 & 14.00 & 24.57 \\
        Qwen3.6-27B           & 23.00 & 30.00 & 30.00 & 16.00 & 38.00 & 18.00 & 12.00 & 23.85 \\
        Qwen3.6-35B-A3B       & 28.00 & 36.00 & 18.00 & 14.00 & 26.00 & 22.00 & 16.00 & 22.85 \\
        \midrule
        \rowcolor{gray!20}
        \multicolumn{9}{c}{\textbf{Open-Source Models (Agentic Tool Use)}} \\
        Qwen3-VL-8B           & 29.00 & 58.00 & 38.00 & 16.00 & 42.00 & 30.00 & 10.00 & 31.85 \\
        Qwen3-VL-30B-A3B      & 31.00 & 58.00 & 34.00 & 14.00 & 50.00 & 22.00 & 16.00 & 32.14 \\
        Qwen3.5-27B           & \underline{54.00} & \underline{78.00} & 54.00 & 36.00 & \underline{68.00} & \textbf{56.00} & \textbf{42.00} & \underline{55.42} \\
        Qwen3.5-35B-A3B       & 42.00 & 56.00 & 42.00 & 18.00 & 38.00 & 22.00 & 20.00 & 34.00 \\
        Qwen3.6-27B           & \underline{54.00} & 74.00 & \textbf{64.00} & \underline{38.00} & 66.00 & 46.00 & 34.00 & 53.71 \\
        Qwen3.6-35B-A3B       & 50.00 & 76.00 & \underline{62.00} & 34.00 & 62.00 & 48.00 & \underline{40.00} & 53.14 \\ \midrule
        \addlinespace[2pt]
        \rowcolor{blue!10}
        \textbf{\Method-8B-SFT} (Ours)       & 39.00 & 58.00 & 44.00 & 30.00 & 54.00 & 32.00 & 18.00 & 39.28 \\
        \rowcolor{blue!10}
        \textbf{\Method-8B-RL} (Ours) & \textbf{56.00} & \textbf{80.00} & \textbf{64.00} & \textbf{40.00} & \textbf{70.00} & \underline{52.00} & \textbf{42.00} & \textbf{57.71} \\
        \bottomrule
    \end{tabular}%
    }
\end{table}

Table~\ref{tab:main_results} compares direct answering with agentic tool use. Proprietary models achieve strong performance even without external tools. Their larger capacity and broad parametric world knowledge allow them to answer many questions directly from the observed video and internal knowledge. In contrast, open-source models perform considerably worse under direct answering, indicating that limited parametric knowledge and passive video observation are insufficient for questions that jointly require precise temporal evidence and external information.

Equipping the open-source models with the same tools as \Method consistently yields substantial improvements over their direct-answer counterparts. This result demonstrates the value of turning video question answering into an active research process that can revisit relevant segments, verify visual entities, retrieve missing knowledge, and inspect source webpages. However, the performance differences among agentic models also show that tool access alone is insufficient. Effective Video Deep Research further requires the model to select and compose tools according to the evolving evidence state rather than invoking them as an unstructured pipeline.

\Method-8B-RL achieves the strongest overall performance among the evaluated open-source agentic models and leads or matches them on most subsets. Despite its 8B backbone, it reaches a level comparable to the strongest proprietary models under direct answering and surpasses newer, larger open-source models operating with the same tools. This result shows that our trajectory construction and task-specific training can compensate for limited model scale by transforming missing parametric knowledge into explicit evidence acquisition. The consistent improvement from \Method-8B-SFT to \Method-8B-RL further confirms that RL strengthens autonomous long-horizon coordination beyond imitation of the synthesized trajectories.

\subsection{Ablation Studies}

Table~\ref{tab:ablation_results} shows that \Method performs best across video lengths and difficulty levels, confirming that each tool contributes to the research process. Removing \texttt{crop\_video} causes a degradation across the benchmark, demonstrating the importance of revisiting candidate segments at higher temporal density and visual resolution before retrieval. Removing \texttt{image\_search} also reduces overall performance, with a more evident effect on long-video subsets. This supports its role in confirming visual entities from localized keyframes and preventing uncertain recognition from propagating into subsequent searches. Among the individual tools, removing \texttt{text\_search} produces the largest overall decline, indicating that many questions require the model to connect video-grounded cues with external background or factual knowledge. The removal of \texttt{visit} further weakens performance, showing that search-result summaries alone are often insufficient and that opening source webpages is necessary to extract detailed evidence. The \texttt{web} ablation jointly removes \texttt{image\_search}, \texttt{text\_search}, and \texttt{visit}, causing a substantially larger degradation than removing any single tool and highlighting their complementarity. Together, these results show how \Method coordinates its tools based on the available evidence: \texttt{crop\_video} grounds the research process in relevant visual evidence, \texttt{image\_search} verifies visual entities, \texttt{text\_search} retrieves missing knowledge, and \texttt{visit} consolidates detailed evidence for the final answer.

\begin{table}[t]
    \centering
    \caption{Ablation results on VideoDR and \Bench. All values are accuracy (\%). Each row removes one tool from \Method-8B-RL, except \texttt{web}, which jointly removes all external retrieval tools. S and L indicate short and long videos, respectively.}
    \label{tab:ablation_results}
    \resizebox{\linewidth}{!}{%
    \begin{tabular}{lcccccccc}
        \toprule
        \multirow{2}{*}{\textbf{Model}} & \multicolumn{1}{c}{\textbf{VideoDR}} & \multicolumn{6}{c}{\textbf{VideoRover-Bench}} & \multirow{2}{*}{\textbf{Avg.}} \\
        \cmidrule(lr){2-2} \cmidrule(lr){3-8}
        & \textbf{Acc.} & \textbf{S-Easy} & \textbf{S-Medium} & \textbf{S-Hard} & \textbf{L-Easy} & \textbf{L-Medium} & \textbf{L-Hard} \\
        \midrule
        \rowcolor{blue!10}
        \Method-8B-RL       & \textbf{56.00} & \textbf{80.00} & \textbf{64.00} & \textbf{40.00} & \textbf{70.00} & \textbf{52.00} & \textbf{42.00} & \textbf{57.71} \\
        $-$ \texttt{crop\_video}         & \underline{45.00} & 74.00 & 52.00 & \underline{34.00} & 64.00 & 42.00 & 36.00 & \underline{49.57} \\
        $-$ \texttt{image\_search}       & 43.00 & \underline{76.00} & \underline{54.00} & \underline{34.00} & 56.00 & 44.00 & 36.00 & 49.00 \\
        $-$ \texttt{text\_search}        & 24.00 & 40.00 & 26.00 & 20.00 & 30.00 & 16.00 & 10.00 & 23.71 \\
        $-$ \texttt{visit}               & 41.00 & \underline{76.00} & 44.00 & 30.00 & \underline{68.00} & \underline{46.00} & \underline{38.00} & 49.00 \\
        $-$ \texttt{web}                 & 13.00 & 16.00 & 18.00 & 14.00 & 16.00 & 10.00 & 10.00 & 13.85 \\
        \bottomrule
    \end{tabular}%
    }
\end{table}

\subsection{RL Training Dynamics}

\begin{figure}[!ht]
  \centering
  \begin{subfigure}{0.45\columnwidth}
    \centering
    \includegraphics[width=\linewidth]{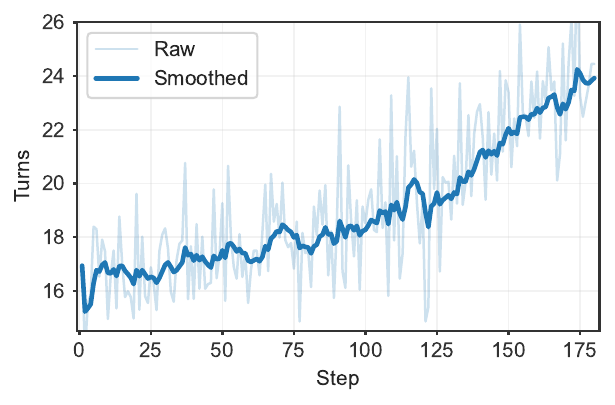}
    \caption{Interaction turns per rollout.}
    \label{fig:num_turns}
  \end{subfigure}
  \hspace{0.05\columnwidth}
  \begin{subfigure}{0.45\columnwidth}
    \centering
    \includegraphics[width=\linewidth]{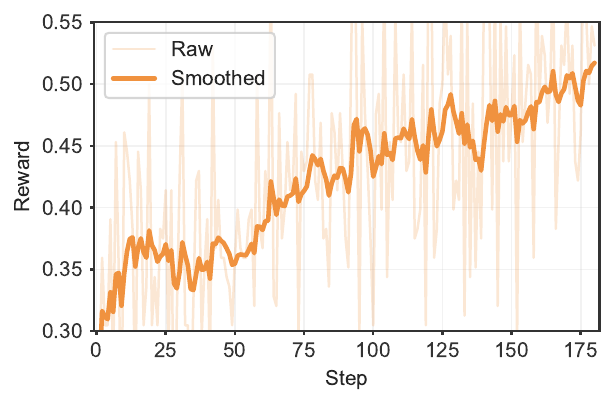}
    \caption{Training reward.}
    \label{fig:training_reward}
  \end{subfigure}
  \caption{RL training dynamics of \Method. Light curves show raw measurements, and dark curves show smoothed trends.}
  \label{fig:training_dynamics}
\end{figure}

As shown in Figure~\ref{fig:training_dynamics}, both the smoothed reward and interaction length increase after the SFT cold start. Because the reward depends on final-answer correctness rather than trajectory length, their joint growth suggests that longer rollouts support sustained evidence acquisition rather than merely adding redundant tool calls. This trend matches the design of our RL data and indicates that RL strengthens the long-horizon coordination of video localization, multimodal search, and webpage browsing. Together with the ablation results, it shows that \Method benefits from learning to compose tools across research trajectories, rather than simply having access to them.

\section{Conclusion}

In this work, we present \Method, a unified framework that iteratively coordinates video observation and open-world retrieval, using each result to guide the next action. To develop and evaluate this capability, we construct verified SFT and RL data covering temporal localization, multimodal search, webpage browsing, and evidence synthesis, together with \Bench across various video durations and research difficulties. Experiments on VideoDR and \Bench show that \Method-8B-RL achieves performance comparable to proprietary models under direct answering while outperforming newer and larger open-source models equipped with the same tools. The ablation results and training dynamics further demonstrate the complementary roles of video grounding and external retrieval, as well as the importance of RL for learning their long-horizon coordination.

\bibliography{main}
\bibliographystyle{main}

\newpage

\appendix
\section{Experimental Details}
\label{app:experimental_details}

\paragraph{Training configuration.}
We use \textit{ms-swift}~\cite{zhao2025swift} for SFT and \textit{VeRL}~\cite{sheng2025hybridflow} for RL. The learning rates are set to $2\times10^{-5}$ and $1\times10^{-6}$, with batch sizes of 256 and 16, for SFT and RL, respectively. During RL, \textit{vLLM}~\cite{kwon2023efficient} samples 8 rollout trajectories per training example, and each rollout can make at most 25 tool calls. For the initial video observation, we sample at 1 FPS, retain at most 256 frames, and limit each frame to a maximum resolution of $224\times224$ pixels. The \texttt{crop\_video} tool instead samples at 2 FPS and returns at most 32 frames at their original resolution. All experiments are conducted on 4 servers, each equipped with 8 NVIDIA H800 GPUs and 2 TB of memory.

\paragraph{Evaluation configuration.}
The proprietary direct-answer baselines comprise GPT-5~\cite{singh2025openai}, GPT-5.2, GPT-5.4, Gemini-2.5-Flash~\cite{comanici2025gemini}, Gemini-2.5-Pro, Gemini-3-Flash~\cite{gemini3}, and Gemini-3-Pro. The open-source baselines comprise Qwen3-VL-8B, Qwen3-VL-30B-A3B, Qwen3.5-27B, Qwen3.5-35B-A3B, Qwen3.6-27B, and Qwen3.6-35B-A3B, each evaluated under both direct answering and ReAct-style agentic tool use~\cite{yao2023react}. In the agentic setting, these baselines, \Method-8B-SFT, and \Method-8B-RL can make at most 50 tool calls per example. For consistent comparison under multimodal input constraints, we cap the initial video input at 64 frames. All other settings follow the training configuration.

\section{Qualitative Cases}

We provide two qualitative examples that illustrate how \Method coordinates video grounding and external retrieval according to the evolving evidence gap. In both cases, the model first localizes and crops a relevant video segment, then uses the observed visual cues to guide subsequent search and webpage verification.

\begin{figure}[t]
    \centering
    \includegraphics[width=\linewidth]{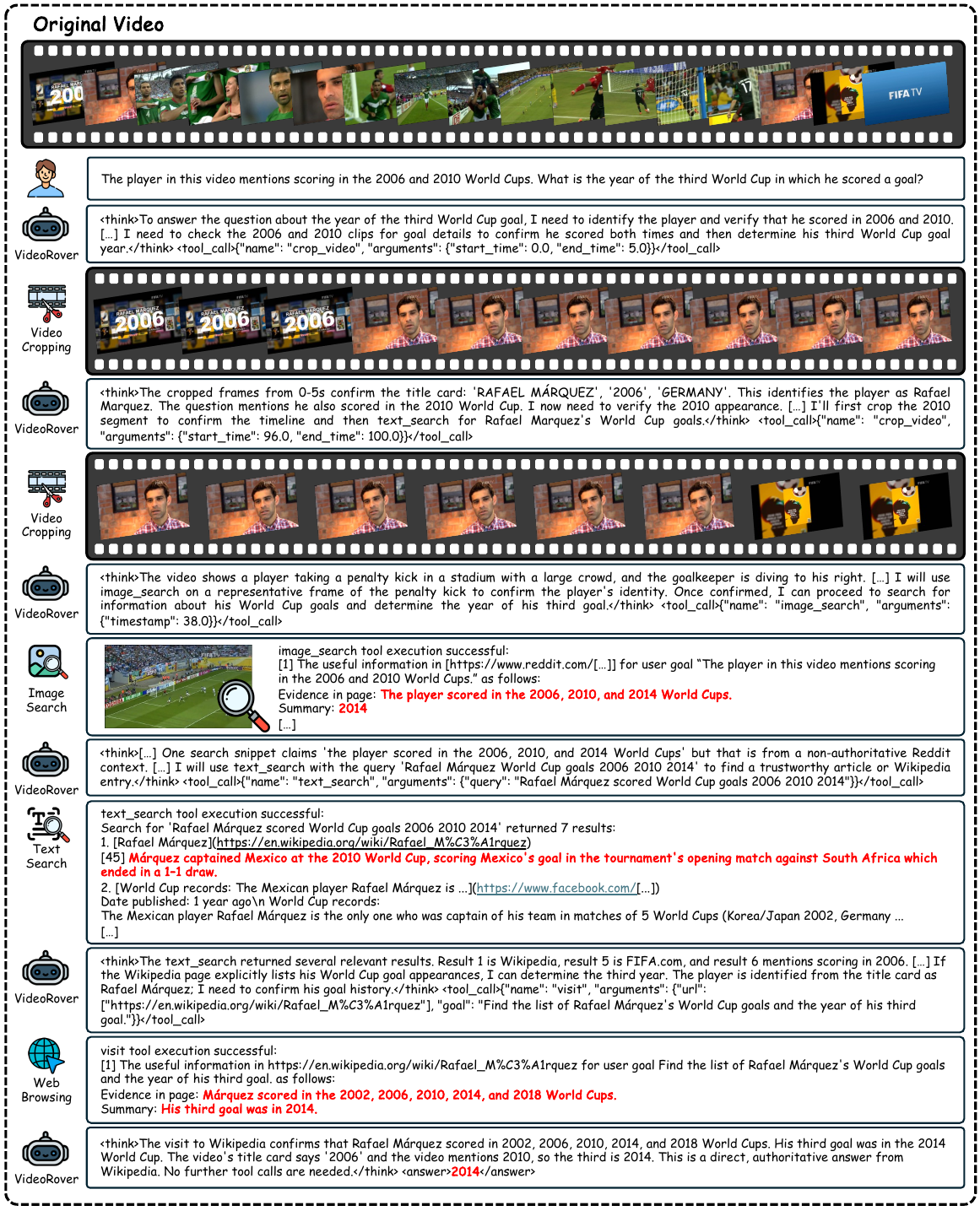}
    \caption{Qualitative case on identifying the year of Rafael M\'arquez's third World Cup goal. \Method crops the interview and match segments, combines image search with text search, and visits a retrieved webpage to verify the final answer.}
    \label{fig:case_1}
\end{figure}

\paragraph{World Cup case.}
Figure~\ref{fig:case_1} shows a trajectory in which the answer requires identifying the player from the video and retrieving his World Cup scoring history. \Method inspects multiple relevant segments, uses image and text search to collect candidate evidence, and visits a source webpage before producing the final answer.

\begin{figure}[t]
    \centering
    \includegraphics[width=\linewidth]{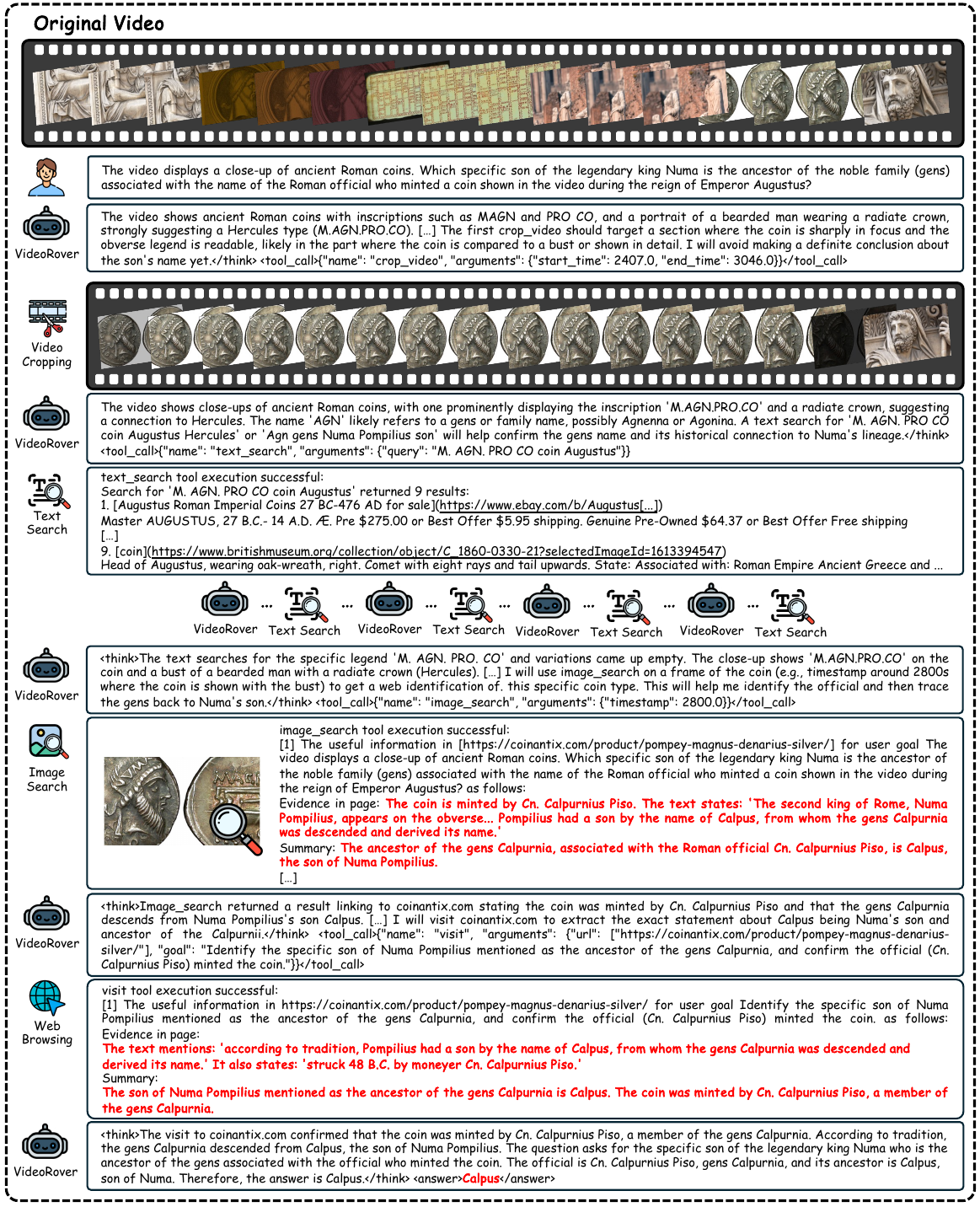}
    \caption{Qualitative case on tracing the ancestry of the gens associated with an ancient Roman coin. After localizing the coin in the video, \Method revises unsuccessful text-search attempts with image search and webpage browsing, ultimately identifying Calpus as the relevant son of Numa Pompilius.}
    \label{fig:case_2}
\end{figure}

\paragraph{Roman coin case.}
Figure~\ref{fig:case_2} presents a more challenging trajectory grounded in a close-up of an ancient Roman coin. When repeated text searches do not identify the relevant gens, \Method switches to image search using the localized keyframe and then visits the retrieved webpage to confirm the historical relationship.

\end{document}